\documentclass[letterpaper,verbose,margin=1in]{article}
 \usepackage[nonatbib, preprint]{neurips_2025}

\workshoptitle{AI4Science: The Reach and Limits of AI for Scientific Discovery}

\usepackage[utf8]{inputenc} 
\usepackage[T1]{fontenc}    
\usepackage{hyperref}       
\usepackage{url}            
\usepackage{booktabs}       
\usepackage{amsfonts}       
\usepackage{nicefrac}       
\usepackage{microtype}      
\usepackage{xcolor}         
\usepackage{graphicx}
\usepackage{amsmath}
\usepackage{enumitem}
\usepackage{subcaption}
\usepackage[linesnumbered,ruled,vlined]{algorithm2e}
\usepackage{multicol}
\setlist[itemize]{leftmargin=*}

\title{The Transparent Earth: A Multimodal Foundation Model for the Earth's Subsurface}

\author{
  Arnab Mazumder\quad
  Javier E. Santos\quad
  Noah Hobbs\quad
  Mohamed Mehana \quad
  Daniel O'Malley\\
  Energy and Natural Resources Security Group\\
  EES-16: Earth and Environmental Sciences Divison\\
  Los Alamos National Laboratory\\
  Corresponding Author Email: \texttt{a\_maz@lanl.gov}
}

\begin{document}

\maketitle





\begin{abstract}
We present the Transparent Earth, a transformer-based architecture for reconstructing subsurface properties from heterogeneous datasets that vary in sparsity, resolution, and modality, where each modality represents a distinct type of observation (e.g., stress angle, mantle temperature, tectonic plate type). The model incorporates positional encodings of observations together with modality encodings,  derived from a text embedding model applied to a description of each modality. This design enables the model to scale to an arbitrary number of modalities, making it straightforward to add new ones not considered in the initial design. We currently include eight modalities spanning directional angles, categorical classes, and continuous properties such as temperature and thickness. These capabilities support in-context learning, enabling the model to generate predictions either with no inputs or with an arbitrary number of additional observations from any subset of modalities. On validation data, this reduces errors in predicting stress angle by more than a factor of three. The proposed architecture is scalable and demonstrates improved performance with increased parameters. Together, these advances make the Transparent Earth an initial foundation model for the Earth's subsurface that ultimately aims to predict any subsurface property anywhere on Earth.
\end{abstract}

\section{Introduction}
Recent advances in machine learning (ML) are reshaping Earth science. ML models has been used for satellite imagery classification \cite{yuan2020deep}, time-series analysis \cite{shreedharan2021machine}, and data augmentation \cite{bonke2024data}. Physics-informed neural networks have also been applied to groundwater flow prediction \cite{ali2024physics} and climate parametrization \cite{alamu2025physics}. However, because observations span diverse modalities and resolutions, many models remain specialized to narrow subdisciplines such as weather forecasting or seismic modeling \cite{sheng2025seismic, bodnar2025foundation}. Foundation models (FMs) show promise for learning from heterogeneous inputs and enabling zero-shot generalization \cite{mai2023opportunities}, yet current Earth-science FMs still target relatively narrow scopes such as seismicity \cite{sheng2025seismic}, climate prediction \cite{bodnar2025a}, or oceanography \cite{bi2024oceangpt}. This persistent specialization reflects the heterogeneity of Earth systems across scales, limited validation data, and the wide variety of data types across fields.

\begin{figure}[ht]
    \centering
    \includegraphics[width=\linewidth]{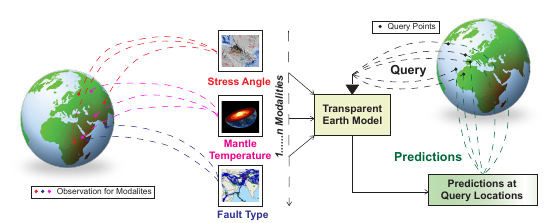}
    \caption{The Transparent Earth workflow fuses heterogeneous observations from multiple modalities and locations to generate predictions for user-defined modalities at arbitrary query points.}
    \label{fig:flow}
\end{figure}

\begin{figure}[h]
    \centering
    \includegraphics[width=\linewidth]{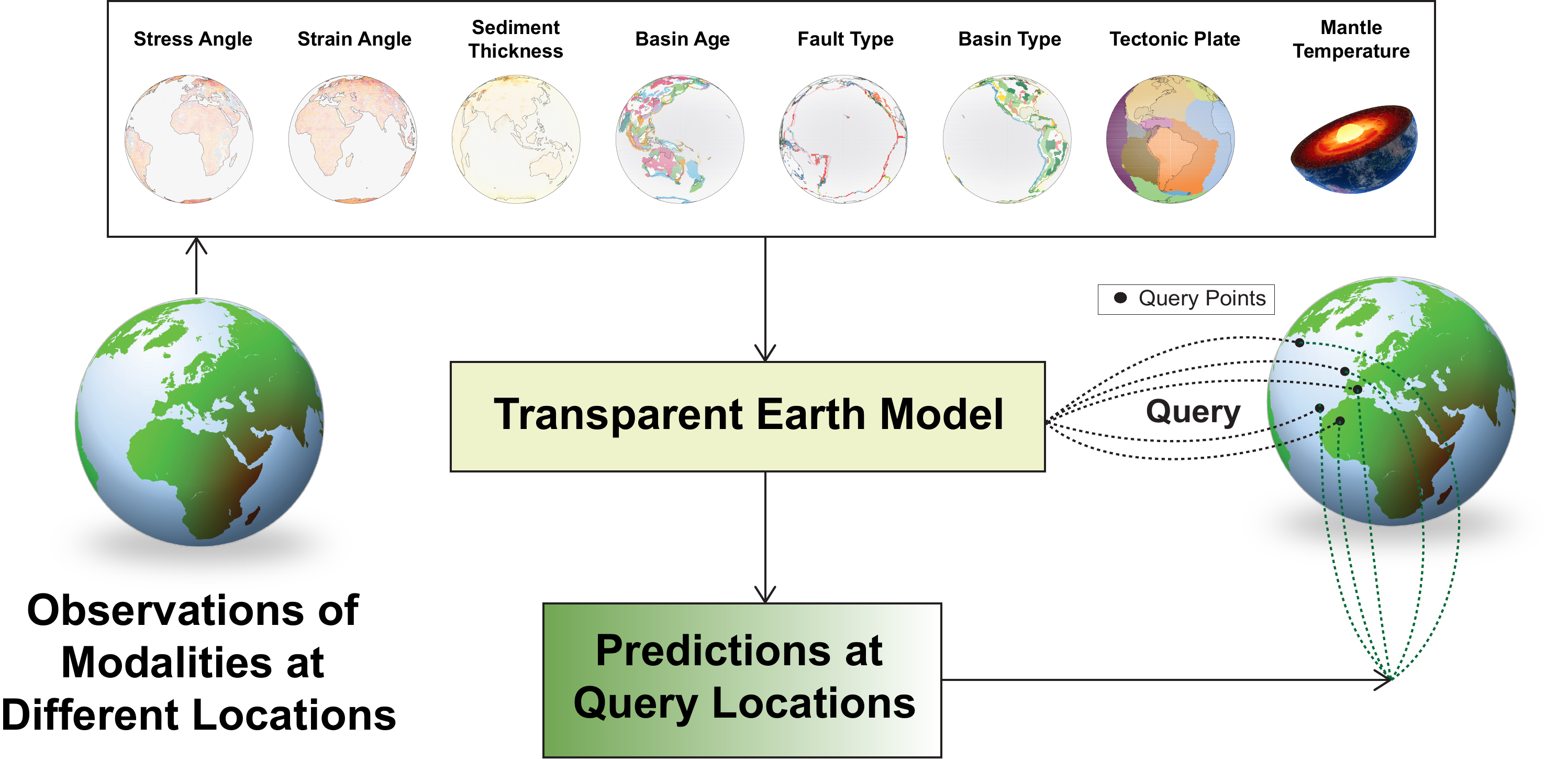}
    \caption{The Transparent Earth workflow showing revolving globes for different modalities to mimic observations coming through to the model from different parts of the world. The eight modalities shown here are the ones that have been used to create the models in this work.}
    \label{fig:flow_v2}
\end{figure}

The integration of multiple modalities across subdisciplines has the potential to substantially improve the resolution and reliability of subsurface property maps. Achieving this, however, requires models that can generalize from sparse and non-uniform observations to provide accurate predictions at local and global scales. To address this challenge, we propose \textit{The Transparent Earth}, a transformer-based framework for global subsurface modeling. As illustrated in Figure~\ref{fig:flow} and \ref{fig:flow_v2} our model learns from sparse, multimodal observations distributed worldwide and reconstructs geophysical fields with  high confidence. Our main contributions are:

\begin{itemize}
\item A unified attention-based architecture that allows directional, categorical, and continuous modalities to cross-attend during encoding and for multitask prediction, covering inputs such as stress angle, tectonic structure, basin attributes, and mantle temperature.
\item A training strategy that handles incomplete, spatially disjoint observations with no overlapping coordinates. Through random sampling of observations  and modalities, the model learns to generalize under real-world sparsity and heterogeneity.
\item A query-driven decoder design that incorporates both spatial and task context, where task embeddings are derived from encoder inputs. This conditioning enables the model to infer \textit{what to predict}  and \textit{where} without relying on memorization.
\end{itemize}

\section{Background}

Traditional ML and deep learning (DL) techniques have seen widespread use in Earth science modeling applications. Conventional methods such as support vector machines (SVM) \cite{koray2023improving}, random forests \cite{misra2019machine}, gradient boosting machines \cite{shahdi2021exploratory}, and shallow neural networks (e.g., multilayer perceptrons) \cite{mao2024efficient} are frequently limited to predictions of individual subsurface attributes. Geophysical foundation models \cite{liu2024foundation} are trained on large datasets such as seismic waveforms to perform multiple tasks such as classification, inversion, and signal processing \cite{sheng2025seismic}. However, these foundation models remain focused on a singular discipline and are unable to leverage information from other modalities (such as subsurface temperature, geology, among others). There has also been significant effort to introduce physics informed models for subsurface engineering, especially for the oil and gas industry \cite{sinha2025review} and energy systems \cite{latrach2024critical}. However, modeling each modality in isolation limits the ability to capture the interconnected nature of subsurface properties. In practice, many of these modalities are inherently correlated. 

For instance, sedimentary basin type, age, and thickness can be correlated to fault types and location that can in turn be correlated to seismic events and ultimately to current stress regimes \cite{laske2013update, evenick2021glimpses}. Furthermore, diverse modalities can inform higher level predictions such as assessments of geothermal prospect viability. These interdependencies, often subtle or non-obvious, can be learned by data-driven models given sufficient observations.

\begin{figure}[h]
    \centering
    \includegraphics[width=\linewidth]{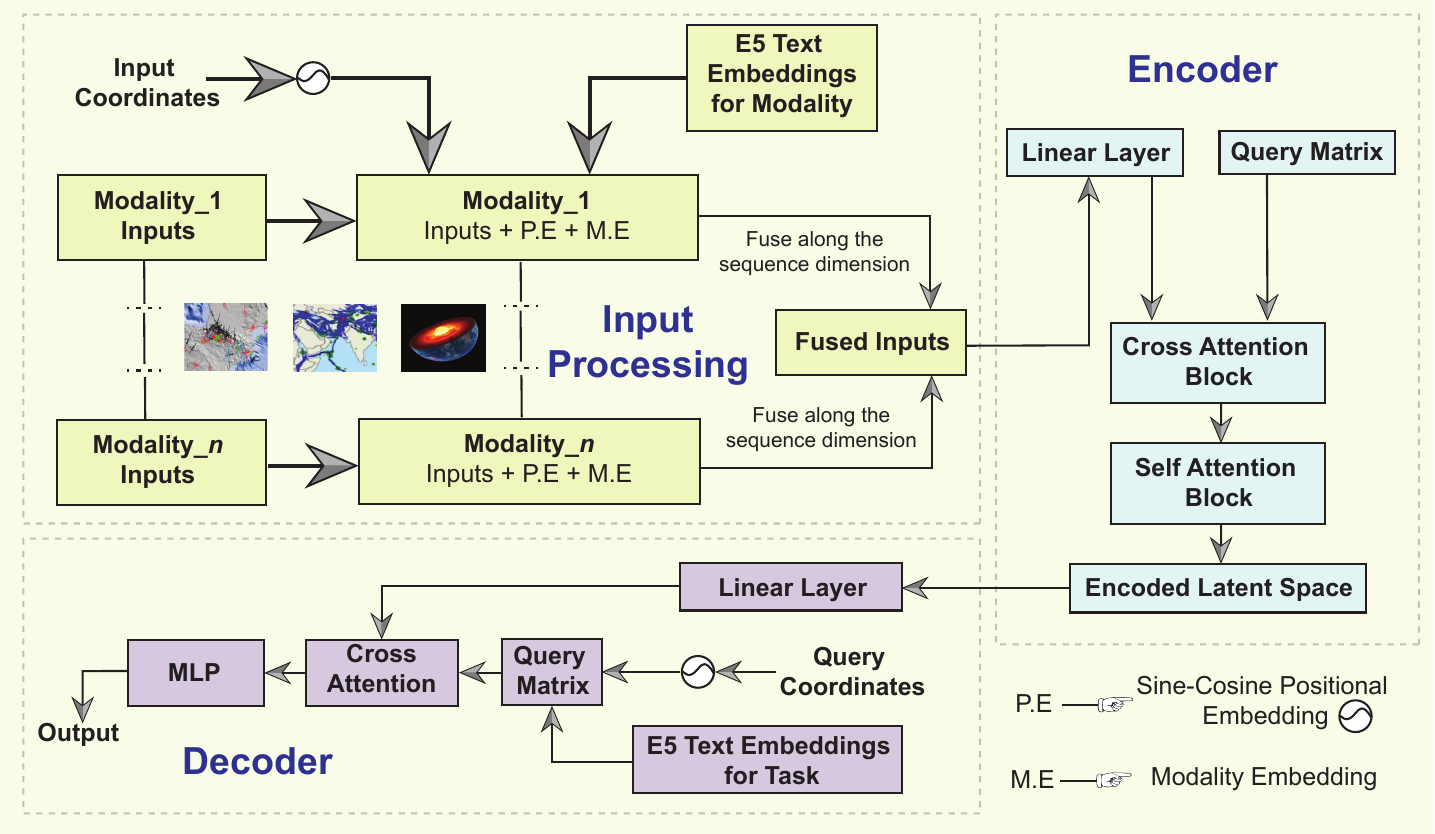}
    \caption{The Transparent Earth architecture for multimodal subsurface modeling. \textbf{(Input Processing)} Inputs are augmented with sine–cosine positional encodings and text-derived modality embeddings, then fused into a unified sequence. \textbf{(Encoder)} The encoder processes these inputs through cross- and self-attention blocks to form a shared latent space. \textbf{(Decoder)} The decoder uses task embeddings and coordinate queries to extract relevant features and generate predictions at arbitrary locations.}
    \label{fig:arch}
\end{figure}

Advancing subsurface property reconstruction and super-resolution requires models that can jointly reason across diverse geophysical modalities spanning multiple geologic disciplines. This is particularly valuable in data-sparse regions \cite{brown2025alphaearth}, where direct observations are scarce, expensive, or highly heterogeneous. Inspired by recent Earth observation systems that leverage globally anchored multimodal fusion architectures \cite{reichstein2019deep, nguyen2023climax, bodnar2025a}, we adopt a similar strategy for subsurface modeling. Specifically, we apply positional encodings of geographic coordinates \cite{vaswani2017attention}, following prior implementations \cite{bodnar2025a, santos2023development}, to correlate data across disciplines within a unified spatial reference frame.

\begin{table}[h]
\centering
\caption{Summary of datasets used in the Transparent Earth. Each modality is labeled with its task type, spatial resolution, value range or number of classes, sparsity, and  source.}
\label{tab:dataset}
\resizebox{\textwidth}{!}{%
\begin{tabular}{lcccccl}
\toprule
\textbf{Modality}             & \textbf{Task Type}    & \textbf{Resolution}             & \textbf{Value Range / Classes}                          & \textbf{Sparsity}       & \textbf{Source}                       \\ \midrule
Stress Angle         & Regression     & $0.5^\circ \times 0.5^\circ$      & $0^{\circ}$ – $180^{\circ}$      & $\checkmark$ & \cite{heidbach2018smoothed}  \\ \midrule
Strain Angle         & Regression     & $0.6^{\circ} \times 0.5^{\circ}$       & $0^{\circ}$ – $180^{\circ}$      & $\checkmark$ & \cite{kreemer2003integrated} \\ \midrule
Sediment Thickness   & Regression     & $1^{\circ} \times 1^{\circ}$           & $0 - 21$ Km                   & $\times$     & \cite{laske2013update}       \\ \midrule
Mantle Temperature  & Regression     & $5^{\circ} \times 5^{\circ}$           & $400^{\circ}$ – $1300^{\circ}$ C & $\checkmark$ & \cite{artemieva2006global}   \\ \midrule
Tectonic Plates    & Classification & $0.001^{\circ} \times 0.001^{\circ}$ & $52$   classes                         & $\times$     &  \cite{bird2003plate}                            \\ \midrule
Fault Type         & Classification & $0.1^{\circ} \times 0.1^{\circ}$       & $24$      classes                      & $\times$     &    \cite{styron2020gem}                          \\ \midrule
Basin Type          & Classification & $0.001^{\circ} \times 0.001^{\circ}$ & $9$    classes                        & $\times$     & \cite{evenick2021glimpses}   \\ \midrule
Basin Age         & Classification & $0.001^{\circ} \times 0.001^{\circ}$ & $17$      classes                      & $\times$     & \cite{evenick2021glimpses}   \\ 
\bottomrule
\end{tabular}%
}
\end{table}

A key component of any multimodal system is the fusion of modality-specific representations \cite{chang2018multisensor}, which can be achieved at different stages of the pipeline. Early fusion combines modalities at the input level \cite{hussain2024comprehensive}, while late fusion integrates their latent representations \cite{chen2024s, gadzicki2020early}. These strategies have been successfully applied across a variety of multimodal tasks \cite{luo2024introducing, xiang2024multimodal, carrillo2022machine} and are equally applicable in the context of geophysical field reconstruction. Leveraging attention-based architectures, such a system could learn cross-modal interactions and spatial dependencies, laying the foundation for a generalized model of the Earth’s subsurface.

\section{Model}

The Transparent Earth model shown in Figure~\ref{fig:arch} is a transformer-based architecture designed to learn subsurface geophysical fields by fusing diverse geospatial modalities. It ingests multiple data sources (e.g., stress angle, strain angle, tectonic plate, fault type, basin type, basin age, sediment thickness, and mantle temperature), each associated with a specific latitude–longitude coordinate, and encodes them into a common latent representation. Each observed input is grounded in space using sine–cosine positional encodings of its coordinates \cite{ramachandran2025primer, santos2023development, jaegle2021perceiver} and augmented with a modality-specific embedding to indicate its data type. The model performs an early fusion of these modality-anchored embeddings, combining all input modalities into a unified sequence that is fed into an attention-based encoder–decoder pipeline. This end-to-end design, shown in Algorithm~\ref{alg:train}, enables the model to capture complex spatial patterns and inter-modal relationships. Finally, the trained model is capable of generating predictions of all modalities at unknown query locations thus generalizing the learned representation of the modalities to novel points on the globe. More details about the data sources is shown in Table~\ref{tab:dataset} and the preprocessing steps can be found in the appendix.

\subsection{Input Processing and Encoder}

Per Algorithm~\ref{alg:train} in the input processing stage, each observation is transformed into a token embedding that encodes both its content and location. Specifically, for each input modality $\mathcal{M}i$, we randomly sample a number of observations $k_i$ drawn from a uniform distribution $\mathcal{U}(1, k{\max})$, as shown in ComputeMod$(k_i)$. This function extracts both the feature vectors $f_i$ (e.g., stress angle, strain angle) and their corresponding spatial coordinates $c_i$.

Each coordinate $c_i$ is passed through a positional encoding function PosEnc$(c_i)$, which applies a sinusoidal transform to encode geographic location. Simultaneously, a modality embedding vector $m_i$ is retrieved via ModEmbed($\mathcal{M}_i$) to uniquely tag each modality. These three elements — raw features, positional encoding, and modality embedding — are concatenated into a single vector per observation as $x_i = [f_i , || , p_i , || , m_i]$.

\begin{algorithm}[ht]
\caption{Training Flow for the Transparent Earth}
\label{alg:train}
\KwIn{Observations $\{\mathcal{M}_i\}$, config parameters}
\KwOut{Trained model parameters}

\begin{multicols}{2}
\textbf{Encoder-side:} \\
Initialize encoder, decoder, optimizer\;
\For{Each modality $\mathcal{M}_i$}{
  sample $k_i \sim \mathcal{U}(1, k_{\max})$ observations\;
  Extract features and coordinates $(f_i, c_i) = \texttt{ComputeMod}(k_i)$\;
  Compute positional encodings $p_i = \texttt{PosEnc}(c_i)$\;
  Retrieve modality embedding $m_i = \texttt{ModEmbed}(\mathcal{M}_i)$\;
  Merge features: $x_i = [f_i \, || \, p_i \, || \, m_i]$\;
  Project: $z_i = \texttt{Proj}(x_i)$\;
}
Fuse modalities across middle dimesion $Z = \texttt{Fuse}_{\left(z_1, z_2, \dots, z_N\right)}$\;
Encode latents: $H = \texttt{Encoder}(Z)$\;

\columnbreak

\textbf{Decoder-side:} \\
\For{Each modality $\mathcal{M}_i$}{
  Sample $k_i^{(q)} \sim \mathcal{U}(1, k_{\text{max}})$ decoder queries\;
  Assign task IDs $t_i$ for each $q_i$\;
  Compute pos encodings $p_i = \texttt{PosEnc}(q_i)$\;
  Get task embeddings $e_i = \texttt{TaskEmbed}(t_i)$\;
  Form queries: $Q_i = [p_i \, || \, e_i]$\;
}
Stack queries $Q = \texttt{Concat}_{\text{query}}\left(Q_1, Q_2, \dots, Q_N\right)$\;

\textbf{Prediction-side:} \\
Predict: $\hat{y} = \texttt{Decoder}(H, Q)$\;
Slice predictions and targets per modality: $(\hat{y}_i, y_i) = \texttt{SliceByMod}(\hat{y}, y)$\;
Compute per-modality losses: $\mathcal{L}_i = \tfrac{1}{N_i}\sum_{j=1}^{N_i}\texttt{Loss}(\hat{y}_{i,j}, y_{i,j})$\;
Aggregate total loss: $\mathcal{L} = \tfrac{1}{M}\sum_{i=1}^{M}\mathcal{L}_i$\;

Backprop and update parameters\;

\end{multicols}
\end{algorithm}

This merged vector is then passed through a projection layer Proj$(x_i)$ that maps it into the shared latent feature space, yielding a modality-specific representation $z_i$. Once all modalities have been processed, their outputs are concatenated along the middle dimension using Fuse$_(z_1, z_2, \dots, z_N)$ to form a unified embedding sequence $Z$. 

All such vectors are concatenated into a single fused sequence, allowing the model to jointly attend to information across different modalities from the outset. The encoder begins by applying a cross-attention operation between this fused input sequence and a set of learnable latent query vectors \cite{jaegle2021perceiver}. In this cross-attention block, the latent queries act as queries attending to the input tokens (which serve as keys and values), extracting an initial integrated representation of the data into a fixed-size latent space. The output is a set of latent embeddings that is subsequently refined through three self-attention layers (each followed by a multi-layer perceptron (MLP)), which capture higher-order interactions among the latent features. These attention mechanisms enable the encoder to learn inter-modal relationships as self-attention can adaptively highlight correlations between observations and across data types \cite{ramachandran2025primer}. The result of the encoder is a compact latent representation that holistically encodes the fused geophysical information from all input modalities.

\subsection{Decoder}

The decoder produces predictions for all modality at user-defined query coordinates using the encoded latent space. For each prediction modality $\mathcal{M}i$, we sample a number of decoder queries $k_i^{(q)}$ from $\mathcal{U}(1, k{\text{max}})$ — these queries correspond to spatial locations where predictions are to be made. Each query point $q_i$ is encoded using the same positional encoding mechanism as the encoder via PosEnc$(q_i)$, producing $p_i$.

A task identifier $t_i$ associated with each modality is used to retrieve a task-specific embedding via TaskEmbed$(t_i)$. The query vector is then formed by concatenating the position and task information as $Q_i = [p_i , || , e_i]$. Together, these form a spatially anchored query vector that encapsulates both the query’s position and the desired prediction type. This is repeated across all prediction modalities, and the resulting queries are combined using Concat$_\text{query}(Q_1, Q_2, \dots, Q_N)$ to form the complete query set $Q$.

The decoder then uses this query vector to attend to the encoder’s latent representation via a multi-head cross-attention mechanism followed four sets of MLP layers. In this cross-attention, the query vector serves as the query, while the latent embeddings provide keys and values – allowing the model to retrieve relevant fused information from the latent space for that specific location.  

Since predictions are generated jointly for all modalities, we extract per-modality predictions and targets using SliceByMod($\hat{y}$, $y$) to obtain $(\hat{y}_i, y_i)$ for each task. Each pair is passed through a loss function Loss($\hat{y}_i$, $y_i$) tailored to the modality, yielding per-modality losses $\mathcal{L}_i$. Finally, these are averaged into a total loss which is used to update model parameters via backpropagation. 

\subsection{Design Considerations}

\begin{figure}[t]
    \centering
    \includegraphics[width=\linewidth]{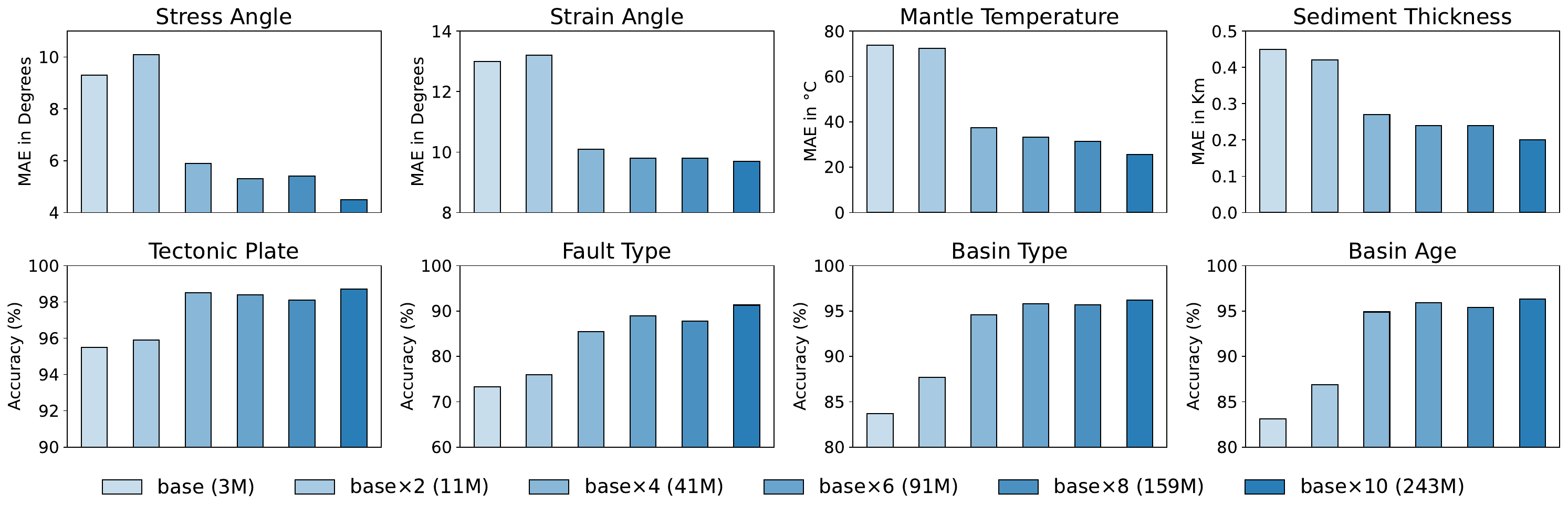}
    \caption{Performance of the Transparent Earth across 8 modalities as model scale increases from 3M to 243M parameters. Regression tasks (top row: stress angle, strain angle, mantle temperature, sediment thickness) are evaluated using mean absolute error (MAE), while classification tasks (bottom row: tectonic plate, fault type, basin type, basin age) are evaluated using accuracy. Scaling up the model lowers regression errors and improves classification results.}
    \label{fig:model_results}
\end{figure}

Our multi-task architecture is designed for geoscientific settings, where each data point may originate from distinct physical processes—such as stress, strain, plate boundaries, faults, or mantle temperature—and is distributed unevenly across the globe. To enable reasoning across these diverse modalities, we incorporate several architectural and data-centric strategies.

\textbf{Fusion along the Sequence Dimension:}
Inputs are fused along the sequence dimension, not the feature dimension, to preserve spatial relationships and retain modality-specific signals. Each input is treated as a location-aware observation, with modality identity encoded via learned embeddings. This avoids over-averaging effects that can arise when fusing heterogeneous features along the feature axis.

\textbf{Dataloader Construction and Randomization:}
The dataloader randomly samples encoder inputs per modality (e.g., 50 stress angle, 30 strain angle, 100 fault type, 200 tectonic plate, 35 mantle temperature) for each batch. Decoder queries, shared across the batch, are geolocations tagged with task identifiers. This setup trains the model to answer both “what” and “where” to predict. Randomizing the number input observations and available modalities promotes robustness to missing data and acts as modality-level dropout, encouraging generalization.

\textbf{Unified Positional Encoding with Depth:}
Mantle temperature inputs vary across depth, unlike surface-only modalities such as stress or plate. To achieve a unified positional encoding across both surface and sub-surface modalities, we apply sinusoidal encoding as shown in Equation~\ref{eq:pos_enc} to normalized latitude ($\phi$) and longitude ($\lambda$) using a fixed set of frequency bands $\{f_i\}_{i=1}^F$ and append the normalized depth value ($z$) directly as an additional dimension. 

\begin{equation}
\mathbf{e} = \left[
\sin(\pi \phi \cdot \mathbf{f}_{\phi}),\ 
\cos(\pi \phi \cdot \mathbf{f}_{\phi}),\ 
\sin(\pi \lambda \cdot \mathbf{f}_{\lambda}),\ 
\cos(\pi \lambda \cdot \mathbf{f}_{\lambda}),\ 
z
\right] \in \mathbb{R}^{4F + 1}
\label{eq:pos_enc}
\end{equation}

This formulation yields a consistent spatial representation $\mathbf{e} \in \mathbb{R}^{4F + 1}$ where $4F$ encodes angular variation across latitude and longitude and the final component $z$ captures vertical positioning. Surface-only modalities are assigned $z = 0$, enabling the model to learn a shared latent space that respects both horizontal and vertical geospatial variation. 

To ensure consistent reconstruction across modalities and resolutions, we selected a fixed sampling frequency of 36 for the latitude direction, with longitude set proportionally to twice this value. This choice is grounded in the Nyquist–Shannon sampling principle, where the maximum frequency $f_{max}$ is determined by the inverse of the grid spacing. For a target reconstruction resolution of $0.5^{\circ} \times 0.5^{\circ}$ in latitude, the corresponding maximum frequency is:

\begin{equation}
f_{\max}^{\text{lat}} = \frac{1}{2 \times 0.5^\circ} = 36, 
\quad 
f_{\max}^{\text{lon}} = 2 \times f_{\max}^{\text{lat}} = 72
\label{eq:nyq_freq}
\end{equation}

This choice in Equation~\ref{eq:nyq_freq} ensures that our encodings faithfully represent the variability present in the datasets while avoiding over-parameterization at unnecessarily finer resolutions.

\textbf{Modality and Task Embeddings:}
Modality and task embeddings are generated using the multilingual E5 text embedding model \cite{wang2024multilingual}. The embeddings are generated by providing a small description about the modality to the text model. In our case, we only used the modality names as description. To prevent modality embeddings from overpowering the positional encodings or observation features after fusion, they are projected to a lower-dimensional space. To that end, we have found that a modality embedding dimension of 8 provides decent balance for providing the modality tag while maintaining the feature information. Task embeddings, in contrast, are projected to match the dimensionality of the query positional embeddings, ensuring compatibility during concatenation.

\textbf{Modality-Specific Loss:} We use loss functions that align with the nature of each prediction task. For angular quantities such as stress and strain angles, we use a specialized angular loss that respects the 180$^\circ$ periodicity of directional data as shown in Equation~\ref{eq:ang_loss}.

\begin{equation}
\mathcal{L}_{\text{angular}} = \frac{1}{N} \sum_{i=1}^{N} \left( \left( \left( \hat{\theta}_i - \theta_i + \frac{R}{2} \right) \bmod R - \frac{R}{2} \right) \bigg/ \frac{R}{2} \right)^2
\label{eq:ang_loss}
\end{equation}

Here, $\hat{\theta}$ and $\theta$ denote predicted and true angles respectively, and $R$ is the angular range (either 180$^\circ$ or 360$^\circ$). The loss $\mathcal{L}_{\text{angular}}$ computes the mean squared angular error after resolving periodicity using modular arithmetic. For classification tasks like tectonic plate type, we use cross-entropy loss and for other regression task we stick to using mean squared error. This task-specific loss design ensures that each modality is optimized in a physically meaningful and statistically appropriate manner.

\section{Results and Analysis}

\begin{figure}[t]
    \centering
    \includegraphics[width=\linewidth]{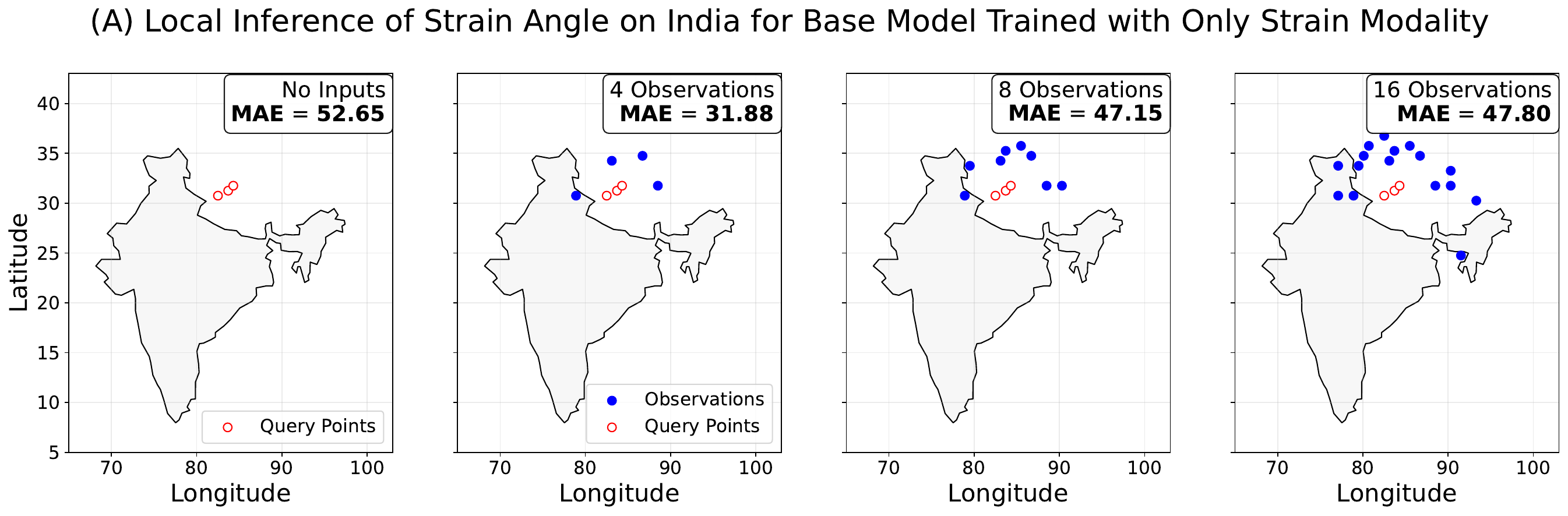}
    \includegraphics[width=\linewidth]{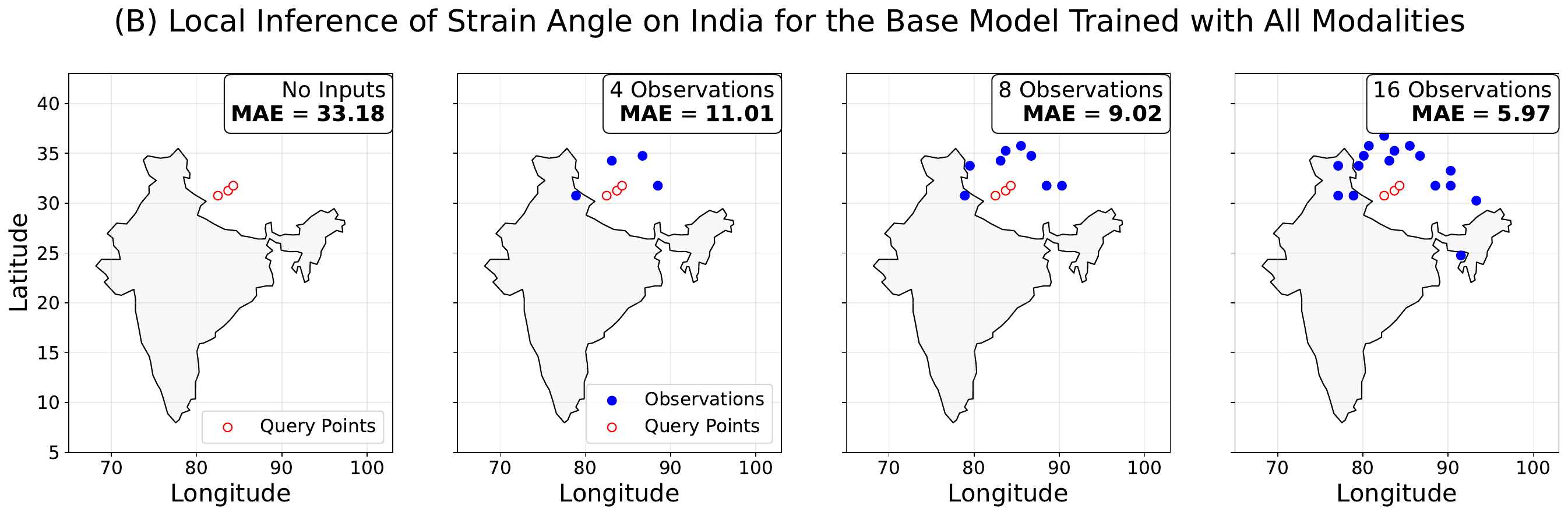}
    \caption{The top four plots shows baseline strain-angle inference along India’s boundaries with the base model trained with strain modality only, while the bottom four show the performance of the same model trained with all modalities for strain prediction. For both tests, With no input observations, the model falls back on global priors, resulting in high error. As nearby observations are added, prediction accuracy progressively improves for the base model trained with all modalities. The model trained with strain modality only shows random trend for MAE with varying observations. Observations and query points are drawn from the held-out test set, with neighbors selected based on proximity to the query point}.
    \label{fig:local_test}
\end{figure}

\begin{figure}[h]
    \centering
    \includegraphics[width=\linewidth]{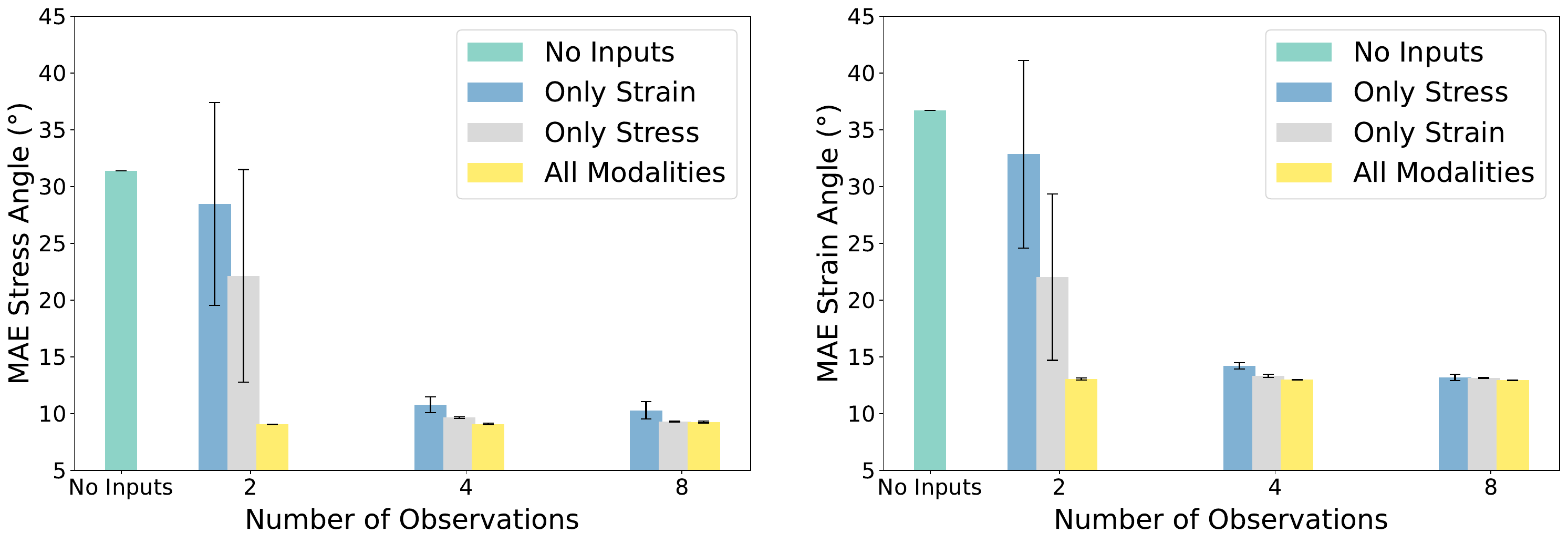}
    \caption{Global inference performance of the baseline model for stress and strain angles with varying numbers of input observations. Bars show MAE when using no inputs, only single-modality inputs (stress or strain), or all modalities. For each case, 2, 4, or 8 observation points are randomly sampled from the training set, and predictions are evaluated on the held-out test set. Error bars indicate standard error across three independent runs.}
    \label{fig:global_test}
\end{figure}

We evaluate the performance of baseline (3M) model up to $10\times$ variants across all regression and classification tasks. All results are reported on the held-out test set, which comprises 5\% of each dataset, while 95\% was used for training. Each model was trained for 200k steps using the Adam optimizer. Figure~\ref{fig:model_results} summarizes performance across model scales and Figures~\ref{fig:local_test} and \ref{fig:global_test} illustrate the behavior of the model for global and local inference with varying observations with our base model configuration. In addition to these plots, detailed difference maps for predictions and ground truths, reconstructed global field maps, local inference test for the continental USA can be found in the appendix.

For the general model performance across all tasks, the encoder latent space was constructed using all modalities. For regression tasks (Figure~\ref{fig:model_results}), MAE decreases consistently with increasing model scale. For classification tasks (Figure~\ref{fig:model_results}), accuracy improves with scale. Tectonic plate classification remains above 95\% across all variants, while fault type prediction shows gradual but steady gains. Basin type and basin age also benefit from scaling, reaching accuracies above 95\% in the largest model.

To illustrate the model’s local inference performance we sample query and observations from the coordinates within the boundaries of India. In this setting, local inference refers to predicting values at query points using only a limited number of nearby observations drawn from the same modality. Here, both the observations and query points are sampled from the held-out strain angle test set using a K-nearest neighbor strategy, and no other modalities are provided to the model. As shown in Figure~\ref{fig:local_test}-(A) and (B), when no input observations are available, the model falls back on global priors, yielding a relatively high error. As nearby strain angle data-points are incrementally added, prediction accuracy improves for the base model trained with all modalities, demonstrating the model’s ability to leverage local context. However, the model trained with only strain modality (Figure~\ref{fig:local_test}-(B)) performs some way off compared to the baseline model trained with all modalities and has random trend for MAE. It struggles to reduce MAE even with high number of observations. These results confirm that the model generalization benefits from localized observations across modalities.

In the global inference setting, the model is asked to predict stress and strain orientation angles for all points in the test set while receiving a limited number of randomly sampled observations from the training set. Performance was averaged across three seed runs with 2, 4, or 8 observations provided. For no inputs case, the model relies on priors, resulting in high errors ($\approx 33^{\circ}$ for stress and $35^{\circ} $for strain). Adding observations from the same modality reduces error and uncertainty, with variance shrinking as the number of data-points increases. The most pronounced improvement is achieved when all modalities are available: even with only two observations, errors drop sharply $(\approx 9^{\circ}$–$13^{\circ})$ and remain consistently low across higher observation counts. This effect arises because with two observations per modality, using all eight modalities yields a total of 16 data-points. The model can therefore leverage information across modalities and produce more reliable predictions. These results highlight that multimodal fusion not only reduces prediction error but also stabilizes performance, while additional datapoints further refine accuracy and reliability in global inference tasks.

\section{Conclusion}

We presented the Transparent Earth, a flexible multimodal transformer framework for subsurface field reconstruction that learns from sparse and heterogeneous observations. Our results demonstrate strong scaling trends, in-context learning capabilities, and consistent gains across regression and classification tasks, highlighting the potential of the Transparent Earth as a foundation model for the Earth’s subsurface. Looking ahead, we aim to broaden the framework by incorporating additional geophysical and geological modalities, and extending it with features of varying spatial resolutions, depth-dependence and coverage. The use of text-derived embeddings opens the door to zero-shot generalization, where natural language descriptions of new modalities could guide predictions without the need of retraining or fine tuning. Together, these developments will enable a more comprehensive model of the Earth that integrates structured and unstructured knowledge, scales with increasing data availability, and supports discoveries across goescientific applications.

\begin{ack}
NH and DO were supported by U.S. Department of Energy, Office of Science (Basic Energy Sciences) Early Career Award number ECA1.
\end{ack}

\bibliographystyle{unsrt}
\bibliography{ref}

\newpage
\appendix

\section{Appendix}

\subsection{Dataset Preprocessing}

The modalities used for training are summarized in Table~\ref{tab:dataset}. Since these datasets exist on different scales, normalization is essential for stable training. In our setup, we apply a simple “divide by max value” criterion, which also facilitates straightforward de-normalization during inference. Sediment thickness is the only modality left unnormalized, as its natural range ($0$–$22$ km) does not introduce bias or instability in convergence. The dense datasets in this work primarily correspond to classification targets, as many are derived from publicly available shapefiles \cite{evenick2021glimpses,bird2003plate}, which enable sampling of an arbitrary number of points from the available classes. After sampling, missing coordinates are filled with the label ‘No Basin’ for basin modalities and ‘None’ for the fault modality.

\subsection{Architecture Details}

The proposed Transparent Earth model is grounded in leveraging attention layers akin to \cite{jaegle2021perceiver} to create structured latent spaces that is used over the decoder. In this design, the decoder is deliberately kept lightweight: it consists of a cross-attention layer followed by four stacked MLPs and a shared output head, as formalized in Equation~\ref{eq:dec_eq} where $x$ is the encoded latent space for the cross attention layer and the output of the preceding layer for all layer afterwards.

\begin{equation}
x \;\mapsto\; \text{CrossAttn}(x) 
          \;\mapsto\; \langle\text{MLP}(x)\rangle {\times 3}
          \;\mapsto\; \text{OutputLayer}(x).
\label{eq:dec_eq}
\end{equation}

The decoder is intentionally designed to remain simple, functioning in a lookup-table–like manner The bulk of the heavy lifting is done by the encoder in terms of creating a generalized embedding space of all the modalities. The encoder in this case follows the following structure of Equation~\ref{eq:enc_eq} where $x$ for the cross attention block is the fused latent space and the $x$ afterwards is the output from the preceding layer:

\begin{equation}
\begin{aligned}
    \text{Cross Attention Block: } &\; x \;\mapsto\; x + \text{CrossAttn}(x) \;\mapsto\; x + \text{MLP}(x), \\
    \text{Self Attention Block: }  &\; x \;\mapsto\; x + \text{SelfAttn}(x) \;\mapsto\; x + \text{MLP}(x), \\
    \text{Encoder Structure: }     &\; \text{CrossAttnBlock} \;\rightarrow\; \left(\text{SelfAttnBlock}\right){\times 3}.
\end{aligned}
\label{eq:enc_eq}
\end{equation}

\begin{table}[h]
\centering
\caption{Model Scaling Details}
\label{tab:scale}
\resizebox{\textwidth}{!}{%
\begin{tabular}{lccccc}
\toprule
\textbf{Model} & \textbf{Channel Width} & \textbf{\# Latent Arrays} & \textbf{\# Cross-Attn Heads} & \textbf{\# Self-Attn Heads} & \textbf{Parameter Count} \\ \midrule
Base           & 256  & 512  & 4  & 2  & 3M   \\ \midrule
Base$\times2$  & 512  & 1024 & 8  & 4  & 11M  \\ \midrule
Base$\times4$  & 1024 & 2048 & 16 & 8  & 41M  \\ \midrule
Base$\times6$  & 1536 & 3072 & 24 & 12 & 91M  \\ \midrule
Base$\times8$  & 2048 & 3072 & 32 & 16 & 159M \\ \midrule
Base$\times10$ & 2560 & 2048 & 40 & 20 & 243M \\ 
\bottomrule
\end{tabular}%
}
\end{table}

The scaled model shown in Figure~\ref{fig:model_results} are variants of the baseline model. In all cases, the encoder and decoder structures defined in Equations~\ref{eq:enc_eq} and \ref{eq:dec_eq} are preserved, while scaling is applied along the channel width, number of attention heads, and number of latent arrays. Here, latent arrays are a fixed set of trainable vectors that serve as a compact bottleneck, repeatedly attending to the inputs to accumulate and represent task-relevant information. For attention heads, we consider the head size to be 64 (as per \cite{vaswani2017attention}) for cross attention and 128 for self attention layers and the number of heads is adjusted accordingly during scaling. Moreover, we try to keep the number of latent arrays $2\times$ of the channel width to allow the model configurations enough expressivity w.r.t the channels except for variants base$\times8$ and base$\times10$. It is important to note that the variants scale linearly for channel width, attention heads and sequence length up to base$\times6$ and for variants beyond base$\times6$, we needed to proportionally change the sequence length w.r.t the channel width so that we could train the model with our allocated GPU memory. The details about the scaling parameters is shown in Table~\ref{tab:scale}.

\subsection{Training Stochasticity}

The training procedure incorporates stochasticity in both the selection of observations per modality and the choice of query points. For encoder inputs, a random number between 1 and 384 determines the number of observations sampled from each modality’s training set at every step. Query points are similarly randomized: given a user-defined upper bound, a random number between 1 and that bound specifies the number of queries per modality. Consequently, both the number of observations and queries vary across steps, encouraging the model to learn under different input conditions. Training proceeds for a large number of steps (200k in our case), ensuring that the model is eventually exposed to nearly the entire dataset. In addition, modalities are randomized at each step, which is implemented in the early fusion process where projected modality inputs are concatenated randomly as shown in Equation~\ref{eq:mod_rand}.

\begin{equation}
\mathcal{I}_{fused} = \operatorname{Concat}\!\left( \{ \ell_i \mid i \in \mathcal{S} \} \right), 
\quad 
\mathcal{S} \sim \text{UniformSubset}\big([0, \ldots, M]\big),
\label{eq:mod_rand}
\end{equation}

Here, $\mathcal{I}_{fused}$ denotes the fused input representation, obtained by concatenating the latent vectors $\ell_i$ corresponding to individual modalities. The index set $\mathcal{S}$ represents the randomly chosen subset of modalities, sampled uniformly from the set of all available modalities $[0, \ldots, M]$, where $M$ is the total number of present modalities at a given training step.

The model is also allowed to train without any inputs during training. The idea is to make the model capable of making predictions only with learned priors. This is incorporated in the early fusion process too. If the chosen number of input modalities to fuse turn out to be 0 in any step, the model then creates a trainable array of all zeros which is passed to the encoder and over time with training this array is updated through backpropagation.

\subsection{Loss Function}

\begin{equation}
\begin{aligned}
\mathcal{L}_{angular}       & : \;\; \text{stress angle, strain angle} \\
\mathcal{L}_{MAE}           & : \;\; \text{mantle temperature, sediment thickness} \\
\mathcal{L}_{CE} & : \;\; \text{tectonic plates, basin type, basin age, fault type} \\
\mathcal{L}_{total}         & : \;\; \frac{1}{N} \sum_{j=1}^{N} \Big( C_1 \cdot \mathcal{L}_{angular}^{(j)} 
                               + C_2 \cdot \mathcal{L}_{MAE\_sed.}^{(j)} 
                               + \mathcal{L}_{MAE\_mantle\_temp}^{(j)}
                               + \mathcal{L}_{CE}^{(j)} \Big)
\end{aligned}
\label{eq:total_loss}
\end{equation}

The loss in our case is driven by modalities. For example, stress angle target predictions need to be routed to angular loss where as samples belonging tectonic plate predictions need to go to categorical cross-entropy. We achieve this by generating task ids during query formation which allows us to slice the respective dimensions from the output layer and then route samples to their specific losses. The training employs three types of losses: angular loss (Equation~\ref{eq:ang_loss}), MAE, and categorical cross-entropy, each reduced using the mean operation to account for the varying number of query points per step. All the respective losses are averaged over to create the total loss which is then used for backpropagation. In our experiments, we have found that the angular losses and mantle temperature MAE loss when scaled by 20 and 10 respectively, balance the total loss and discourage other modalities to dominate the total loss function. Hence, we use 20 and 10 for $C_1$ and $C_2$ .The total loss function is shown in Equation~\ref{eq:total_loss} where $N$ is the total number of queries per step.

\newpage

\subsection{Additional Results}

We illustrate additional plots to provide credence to the model performance. The plots include local inference for the USA, training based on only one modality, difference plots, and reconstructed fields for modalities.

\subsubsection{Local Inference across the Continental USA}

\begin{figure}[h]
    \centering
    \includegraphics[width=\linewidth]{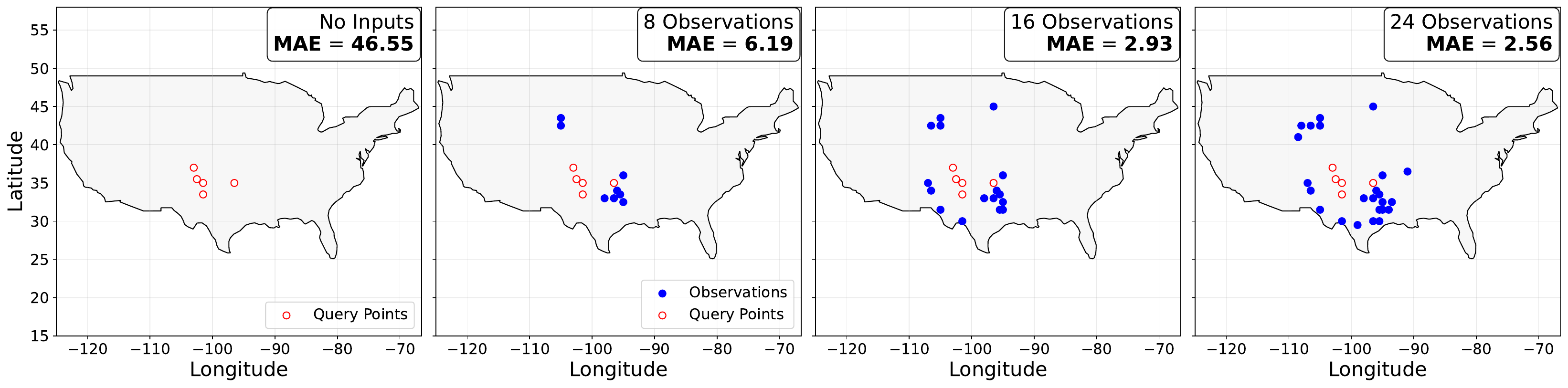}
    \caption{Local inference trend of the baseline model for stress angle along the boundaries of the USA. The observations and queries here are sampled from the held out test set for stress angle using K-nearest neighbors.}
    \label{fig:local_test_usa}
\end{figure}

The local inference test is further extended to show results on the boundaries of the continental USA. The results are obtained using the baseline model and using only the stress observations to create encoded latent space. When no inputs are provided, the model falls back to leveraging the learned priors to make the stress predictions signified by the high MAE value. As more nearby observations are added the model performance gets better progressively and slightly reaches a saturation point between 16 to 24 observations. To select nearby observations, we use the geographic center of the USA as a reference point and perform a brute-force K-nearest neighbor search on the held-out stress test samples relative to this location. This allows us to rank all the points through the euclidean distance w.r.t the reference point. The first five point are then used as queries and the rest are used as encoder observations.

\subsubsection{Difference Plots}

Here, we show the difference plots for all of our tasks except for mantle temperature using the base$\times6$ model. We intentionally, do not show the plot for mantle temperature as it would require showing the query predictions through the Earth at different depths and would be rather difficult to visualize. Instead we encourage the reader to refer to Figure~\ref{fig:model_results} to get an understanding of the model performance on the held out test set for mantle temperature prediction. 

\begin{figure}[h]
    \centering
    \includegraphics[width=\linewidth]{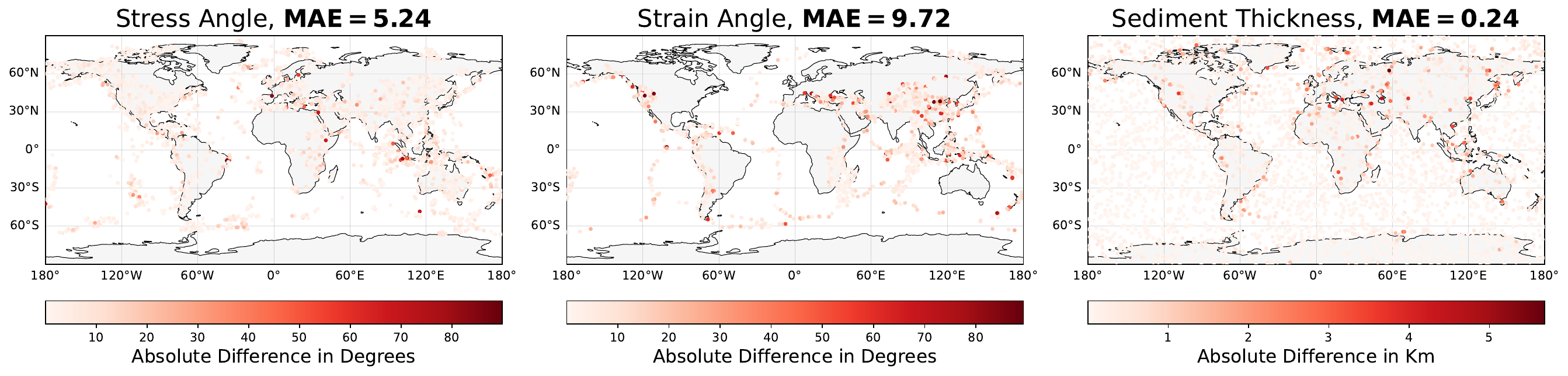}
    \caption{Difference plots for three regression targets using the base$\times6$ model. The data shown here is from the held out test set for respective modalities (5\% of the total set per modality).}
    \label{fig:diff_plot_reg}
\end{figure}

For the regression targets, we show the difference using colormaps. The three plots in Figure~\ref{fig:diff_plot_reg} contains predictions for all the query of testing samples. These data points are sampled from the same dataset where training data is sampled from. It is 5\% of the total points in each modality dataset. Majority of the query predictions are very close to the ground truth as highlighted by the light red color. Any point with dark red corresponds to a point that is off by some large value from the ground truth.

\begin{figure}[h]
    \centering
    \includegraphics[width=\linewidth]{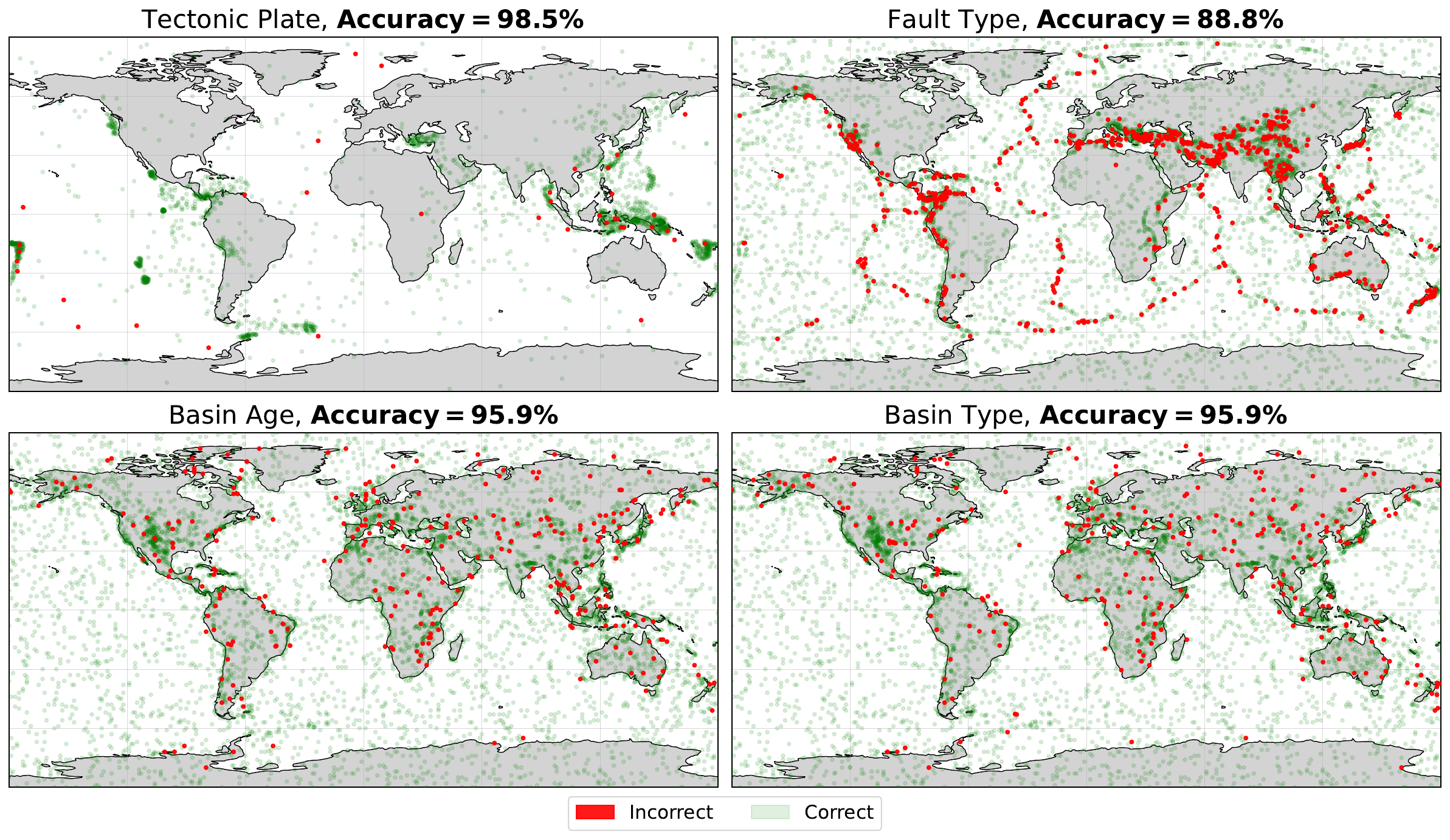}
    \caption{Correctness plots for all classification targets using the base$\times6$ model. The data shown here is from the held out test set for respective modalities (5\% of the total set per modality).}
    \label{fig:diff_plot_cls}
\end{figure}

In a similar vein, for the classification targets, the policy to create testing samples is the same. Any point shown as red on the Figure~\ref{fig:diff_plot_cls} is an incorrect prediction. Particularly, for the fault modality the model struggles. Our understanding is faults are rather difficult to predict due to the random nature of their occurrence along with the fact that the current modalities do not inform the fault modality greatly. As more subsurface properties (i.e seismic velocities, rock types etc.) are introduced fault predictions should improve. 

\subsubsection{Training with Only One Modality}

To further evaluate the impact of multimodal information on generalization, we also trained a model with the same configuration and training settings as the baseline but restricted to a single modality, using only strain data. We then run the same local inference test on the USA boundary along with a global test similar to the one shown in Figure~\ref{fig:global_test}.

\begin{figure}[h]
    \centering
    \includegraphics[width=\linewidth]{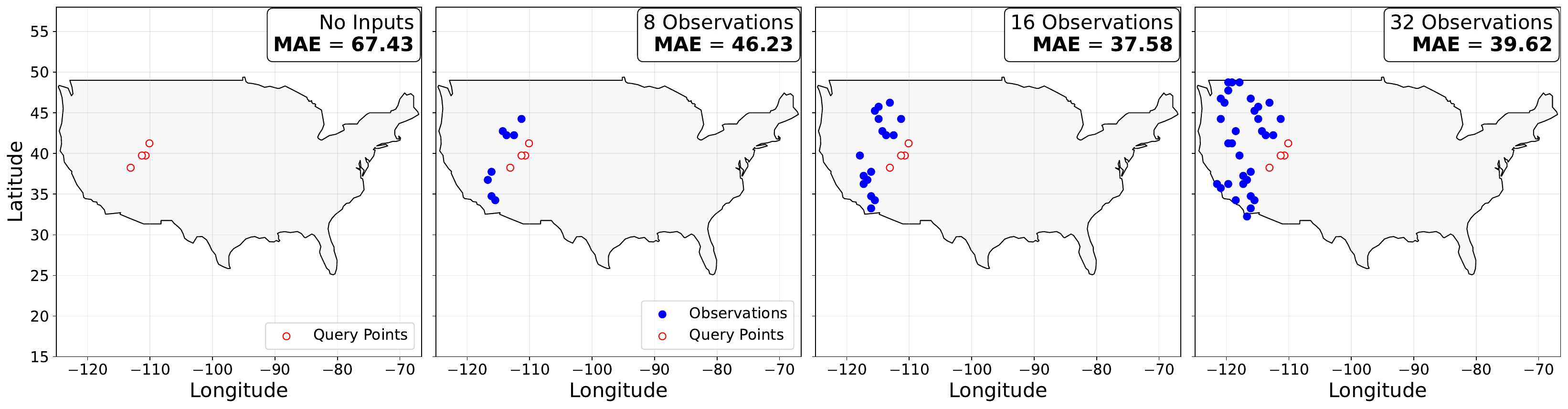}
    \caption{Local inference trend of the baseline model (trained with only strain data) for strain angle along the boundaries of the USA. The observations and queries here are sampled from the held out test set for strain angle using K-nearest neighbors.}
    \label{fig:local_test_usa_strain_only}
\end{figure}

\begin{figure}[h]
    \centering
    \includegraphics[width=0.5\linewidth]{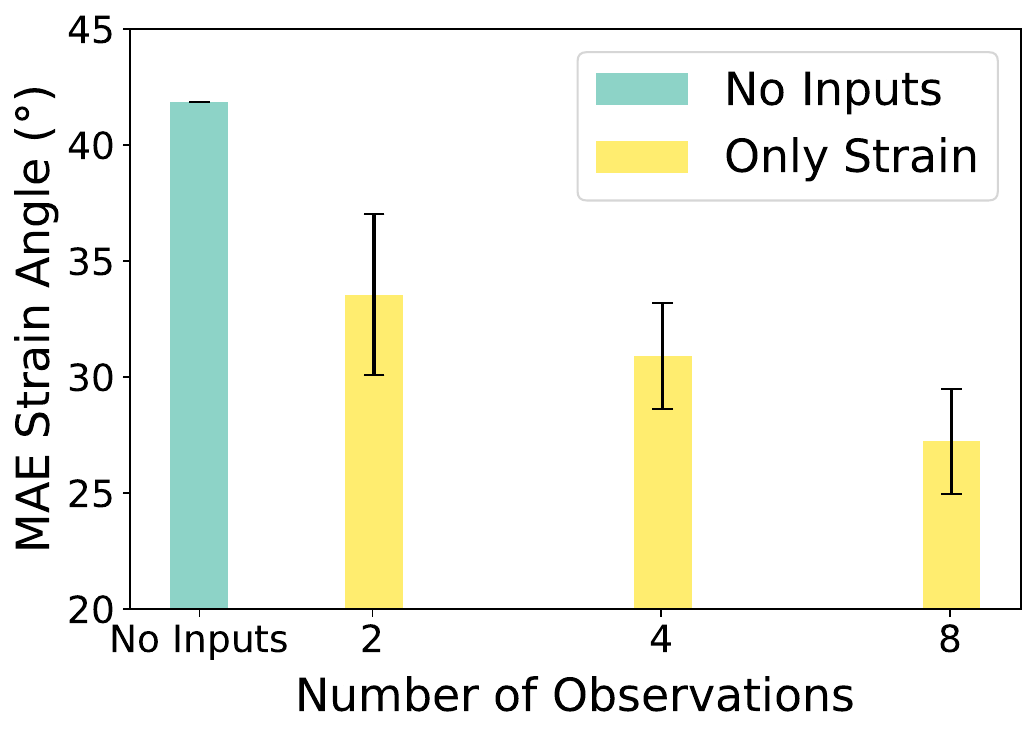}
    \caption{Global inference of the baseline model (trained with only strain data) across different number of strain observations.}
    \label{fig:global_test_strain-only}
\end{figure}

The local inference test (Figure~\ref{fig:local_test_usa_strain_only}) for this setting demonstrates that the model performance drops significantly when no strain inputs are provided. It improves to $37^{\circ}-46^{\circ}$ MAE when strain observations are introduced but still shows considerable error. This makes it evident that different modalities inform each other and allows model to learn features in a more rich fashion. Similarly, for the global test (Figure~\ref{fig:global_test_strain-only} also, the MAE for strain remains more than $25^{\circ}$ (even with 8 observations) which is 90\% more than the MAE of $13^{\circ}$ for the same setting when trained with all modalities.

\subsubsection{Reconstructed Fields}

\begin{figure}[t]
    \centering
    \includegraphics[width=\linewidth]{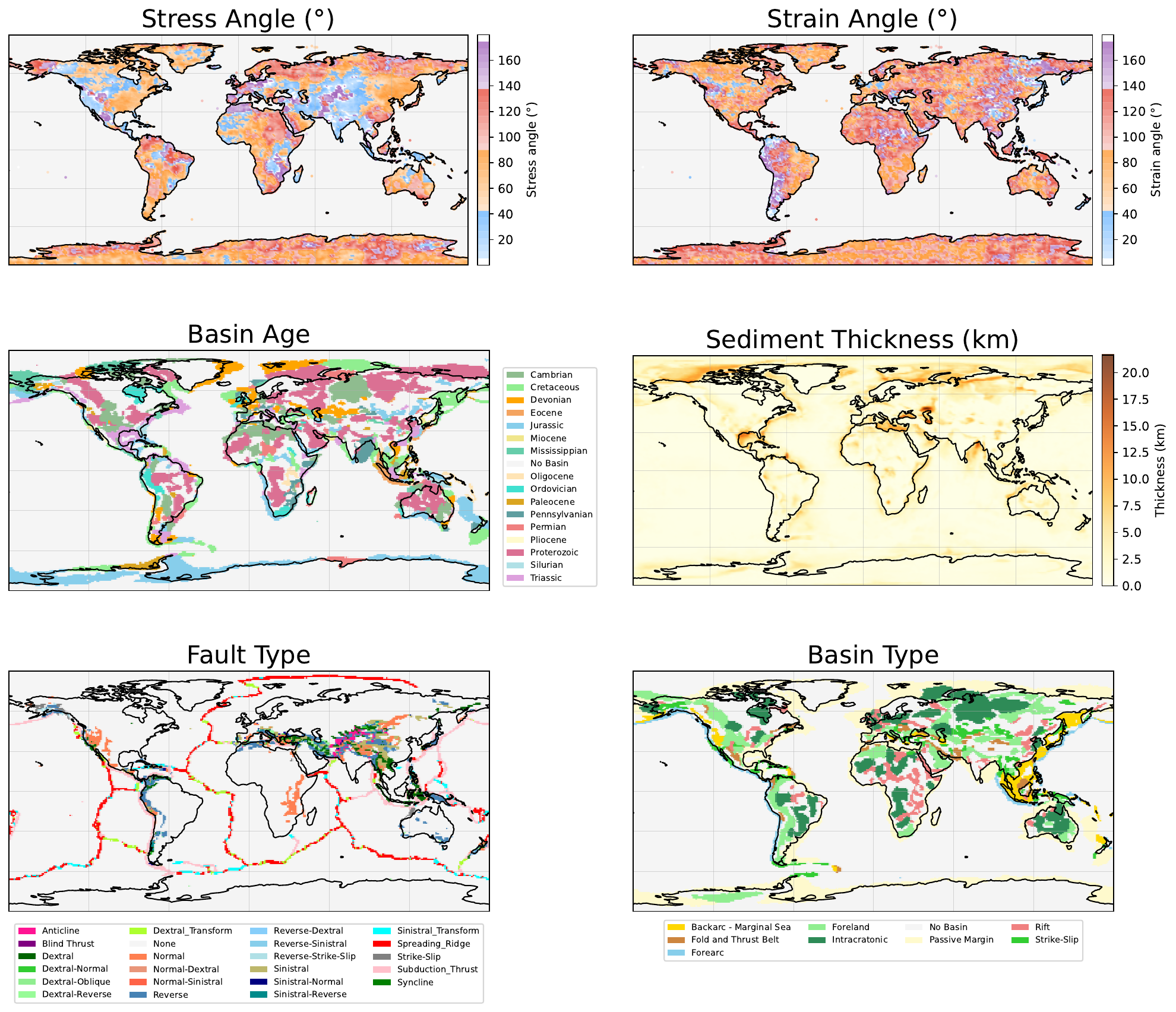}
    \caption{Reconstructed fields on $1\times1$ grid for six different modalities with the base$\times6$ model.}
    \label{fig:field_plots}
\end{figure}

We also show the reconstructed fields on an $1\times1$ grid using the base$\times6$ model. These fields provide an insight into how the model can map different modalities on the global scale. Even though we show the fields on an $1\times1$ grid, the model is capable of creating predictions across any user defined resolution. For the stress and strain angle field we do not show the predictions for the oceans as there are no ground truth for stress or strain angles measured over oceans. For basin age and basin type, the reconstructed fields are a 95\% match to the field shown in \cite{evenick2021glimpses}. Similarly the reconstructed sediment thickness field is has a 0.3~Km MAE from the one created through the crust1.0 model \cite{laske2013update}.

\end{document}